\documentclass[conference]{IEEEtran}
\IEEEoverridecommandlockouts
\usepackage{cite}
\usepackage{amsmath,amssymb,amsfonts}
\usepackage{graphicx}
\usepackage{multirow}
\usepackage{textcomp}
\usepackage{tabularx} 
\usepackage{array}    
\usepackage[margin=1in]{geometry}
\usepackage{booktabs} 
\usepackage{adjustbox} 
\usepackage[table]{xcolor} 
\usepackage{algorithm}
\usepackage{algpseudocode}
\usepackage{subcaption}

\def\BibTeX{{\rm B\kern-.05em{\sc i\kern-.025em b}\kern-.08em
    T\kern-.1667em\lower.7ex\hbox{E}\kern-.125emX}}
\begin{document}

\title{ 
Shedding Light on Depth: Explainability Assessment in Monocular Depth Estimation
}



\author{\IEEEauthorblockN{Anonymous Authors}}

\author{
    \IEEEauthorblockN{
        \begin{tabular}{ccc}
            \begin{tabular}{c}
                1\textsuperscript{st} Lorenzo Cirillo \\
                \textit{Sapienza University of Rome} \\
                Rome, Italy \\
                lorenzo.cirillo@uniroma1.it
            \end{tabular} &
            \begin{tabular}{c}
                2\textsuperscript{nd} Claudio Schiavella \\
                \textit{Sapienza University of Rome} \\
                Rome, Italy \\
                claudio.schiavella@uniroma1.it
            \end{tabular} &
            \begin{tabular}{c}
                3\textsuperscript{rd} Lorenzo Papa \\
                \textit{Sapienza University of Rome} \\
                Rome, Italy \\
                lorenzo.papa@uniroma1.it
            \end{tabular}
        \end{tabular}
    }
    \vspace{0.3cm}
    \IEEEauthorblockN{
        \begin{tabular}{cc}
            \begin{tabular}{c}
                4\textsuperscript{th} Paolo Russo \\
                \textit{Sapienza University of Rome} \\
                Rome, Italy \\
                paolo.russo@uniroma1.it
            \end{tabular} &
            \begin{tabular}{c}
                5\textsuperscript{th} Irene Amerini \\
                \textit{Sapienza University of Rome} \\
                Rome, Italy \\
                irene.amerini@uniroma1.it
            \end{tabular}
        \end{tabular}
    }
}

\maketitle

\begin{abstract}
Explainable artificial intelligence is increasingly employed to understand the decision-making process of deep learning models and create trustworthiness in their adoption. However, the explainability of Monocular Depth Estimation (MDE) remains largely unexplored despite its wide deployment in real-world applications. 
In this work, we study how to analyze MDE networks to map the input image to the predicted depth map.
More in detail, we investigate well-established feature attribution methods, Saliency Maps, Integrated Gradients, and Attention Rollout on different computationally complex models for MDE: METER, a lightweight network, and PixelFormer, a deep network. We assess the quality of the generated visual explanations by selectively perturbing the most relevant and irrelevant pixels, as identified by the explainability methods, and analyzing the impact of these perturbations on the model's output.
Moreover, 
since existing evaluation metrics can have some limitations in measuring the validity of visual explanations for MDE,
we additionally introduce the Attribution Fidelity. This metric evaluates the reliability of the feature attributions by assessing their consistency with the predicted depth map.
Experimental results demonstrate that Saliency Maps and Integrated Gradients have good performance in highlighting the most important input features for MDE lightweight and deep models, respectively. Furthermore, we show that Attribution Fidelity effectively identifies whether an explainability method fails to produce reliable visual maps, even in scenarios where conventional metrics might suggest satisfactory results.
\end{abstract}

\begin{IEEEkeywords}
Explainable AI Evaluation, Model Transparency, Monocular Depth Estimation, Input Feature Attribution
\end{IEEEkeywords}

\section{Introduction}
Monocular Depth Estimation (MDE) \cite{mde} is a computer vision task that has achieved a lot of interest in the last few years, due to its employment in many applications such as robotics and autonomous driving \cite{robotics}. 
In its most general form, MDE predicts a dense depth map from a single input RGB image. Since depth cannot be determined through triangulation in this case, the algorithm must infer depth based on the scene, surrounding objects, and contextual information. In these scenarios, reliable real-time depth predictions are crucial to ensure both the efficiency and trustworthiness of such technologies.
Nowadays, MDE is addressed through deep learning models, particularly Convolutional Neural Networks (CNNs) \cite{b1} \cite{speed}, and Vision Transformers (ViTs) \cite{mde_vit} \cite{pixelformer}. 
In general, CNNs can reach some good results in terms of computational velocity, while ViTs leverage the attention mechanism to achieve high accuracy.
To reach an optimal trade-off between accuracy and efficiency, some lightweight models have also been introduced to perform MDE on hardware-constrained devices \cite{meta_pyra_meter} \cite{meter}. 

Although artificial intelligence has been employed to predict depth in recent years, the explainability of MDE models remains a largely unexplored field \cite{mde_xai_1} \cite{mde_xai_2}, despite its importance in ensuring trust and reliability in real-world contexts (e.g. self-driving cars and healthcare) where small errors can lead to dramatic consequences.

Generally speaking, eXplainable Artificial Intelligence (XAI) is an increasing field of study that aims to understand black-box models' decision-making process, providing explanations to this end. In computer vision, such explanations usually take the form of visual maps, such as saliency maps or heatmaps \cite{xai_vit_survey}. Regarding input feature attribution methods that aim to attribute a relevance score to each input feature, these maps highlight the regions of the input image that most contribute to the output. Previous research on explainability for vision focuses mainly on tasks such as image classification \cite{saliency} and generation \cite{img_gen_xai}. As a consequence, the study of the behaviour of models for dense tasks, such as MDE, has not been addressed in detail due to the intrinsic complexity of the task itself. Unlike image classification, where most of the feature information is concentrated in the class token that can be exploited to obtain the important regions of the input \cite{xai_vit_survey}, MDE involves pixel-wise predictions, where each pixel contributes to the prediction of the others. Hence, the feature information is sparse over the input and the model, making the task more challenging to interpret. Furthermore, evaluating such visual explanations is essential to assess their effectiveness and compare various approaches, but at the same time challenging due to the difficulty in analyzing the effective correspondence with the model prediction.
Existing metrics faithfully evaluate explainability methods in classification scenarios, but can have some limitations in the MDE context.

\begin{figure*}[h]
    \centering
    \setlength{\tabcolsep}{1pt} 
    \renewcommand{\arraystretch}{1} 
    \begin{tabular}{cccccc}
        \adjustbox{valign=m,rotate=90}{METER} &
        \includegraphics[width=0.17\textwidth]{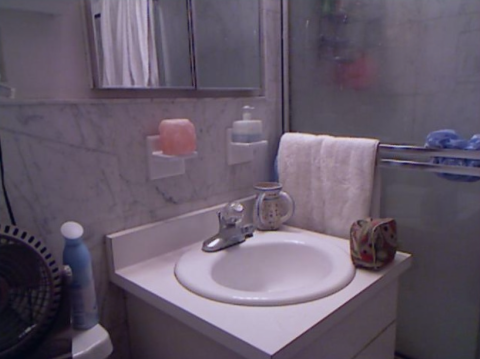} &
        \includegraphics[width=0.17\textwidth]{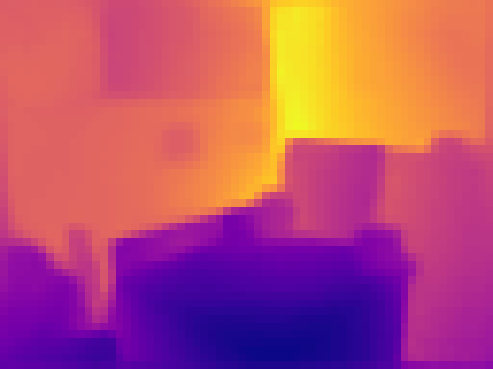} &
        \includegraphics[width=0.17\textwidth]{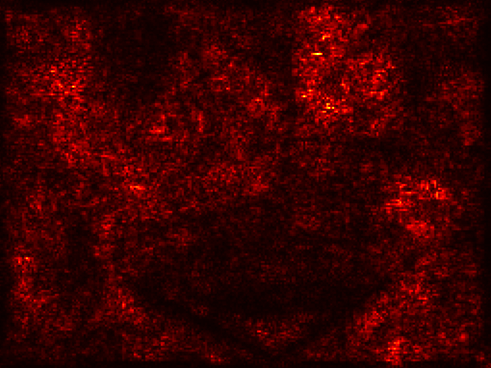} &
        \includegraphics[width=0.17\textwidth]{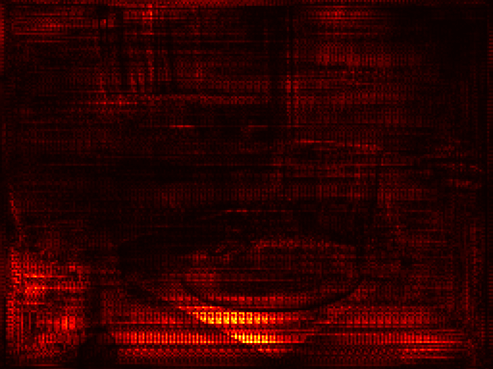} &
        \includegraphics[width=0.17\textwidth]{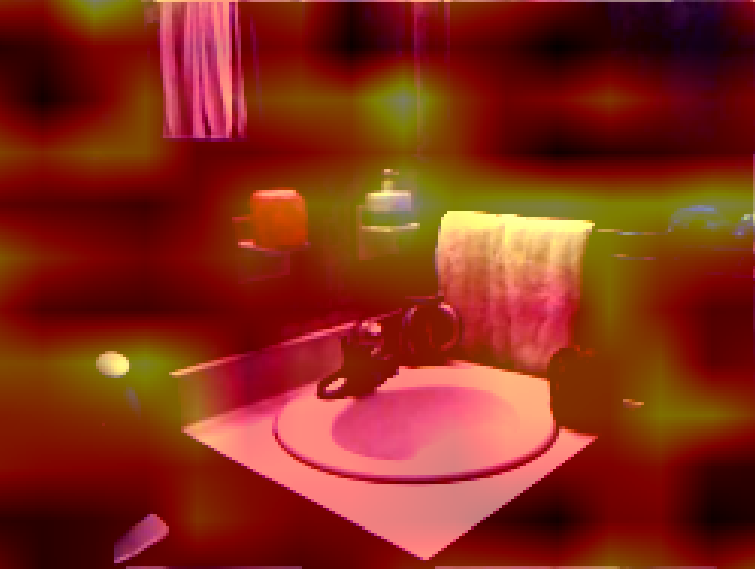} \\
        \adjustbox{valign=m,rotate=90}{PixelFormer} &
        \includegraphics[width=0.17\textwidth]{imgs/input.png} &
        \includegraphics[width=0.17\textwidth]{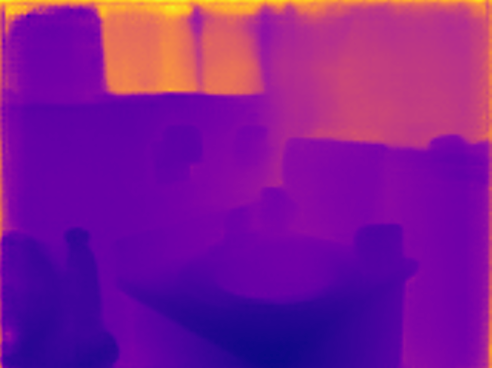} &
        \includegraphics[width=0.17\textwidth]{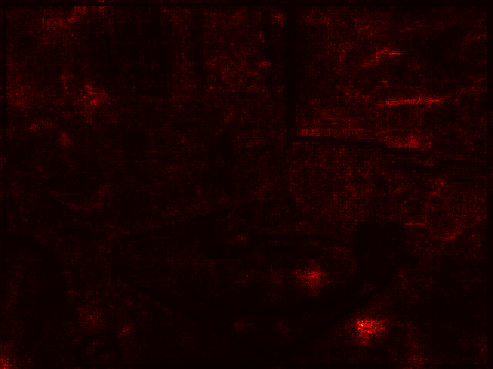} &
        \includegraphics[width=0.17\textwidth]{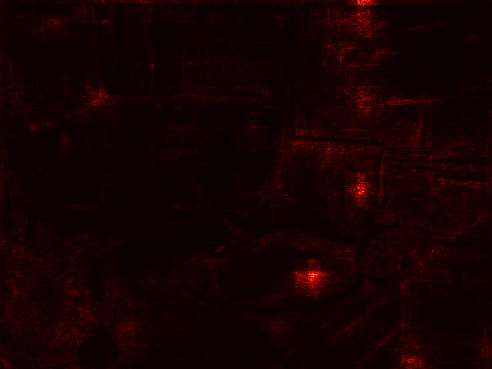} &
        \includegraphics[width=0.17\textwidth]{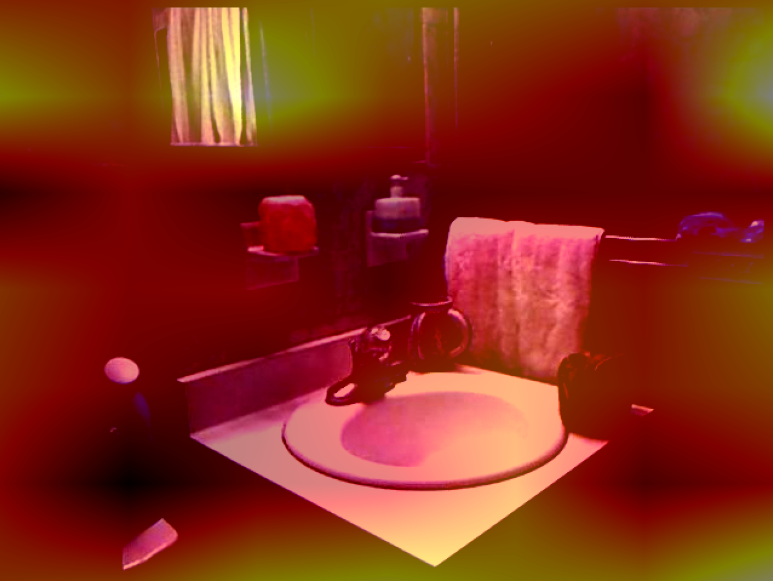} \\
        & (a) & (b) & (c) & (d) & (e) \\
    \end{tabular}
    \caption{Graphic comparison of the generated visual explanations for METER (first row) and PixelFormer (second row). (a) Input image. (b) Predicted depth map. (c) Saliency Map explanation. (d) Integrated Gradients explanation. (e) Attention Rollout explanation.}
    \label{fig:visual_maps_examples}
\end{figure*}

Therefore, in this paper, we conduct a study on the explainability of MDE models to enhance their trustworthiness in real-world scenarios, aiming to push the boundaries of XAI applied to MDE.
Specifically, we analyze three well-established methods from the literature: Saliency Maps \cite{saliency}, Integrated Gradients \cite{int_grad} and Attention Rollout \cite{attn_roll}. Saliency Maps and Integrated Gradients are largely adopted gradient-based methods that can be applied to any model by accessing its gradient. Attention Rollout, on the other hand, exploits attention scores to aggregate contributions across multiple layers of transformer-based architectures. We evaluate these methods on two representative MDE models: PixelFormer \cite{pixelformer}, a deep ViT-based architecture, and METER \cite{meter}, a lightweight hybrid ViT for efficient inference on resource-constrained devices. By considering both deep and lightweight models, we provide a broader perspective on the effectiveness of these explainability techniques across different model complexities. Fig. \ref{fig:visual_maps_examples} depicts some examples of generated visual explanations by the mentioned three explainability methods for METER and PixelFormer. 
We color mapped the visual explanations by assigning yellow color to the most relevant pixels, red color to the somewhat relevant pixels, and black color to the least relevant pixels.
Furthermore, to address the limitations of existing evaluation approaches in the MDE context, we introduce a novel metric called Attribution Fidelity (AF). Such a metric comprehensively evaluates the explanations, enabling a more precise analysis of how well the visual maps correspond to the model prediction and facilitating the comparison between different explainability techniques. 

Experimental results show the effectiveness of the AF metric in evaluating and comparing explainability methods, asserting the gradient-based techniques as reliable tools for explaining the model output.


In summary, the contributions of our work are the following:
\begin{itemize}
    \item A study of explainability of MDE models by adopting three well-established input feature attributions methods from the literature: Saliency Maps, Integrated Gradients, and Attention Rollout.
    \item An evaluation of these methods on two state-of-the-art models for MDE: PixelFormer and METER, representing a deep and lightweight architecture, respectively.
    \item The proposal of a novel evaluation metric, AF, to comprehensively assess the generated visual explanations in the MDE scenario.
\end{itemize}

The rest of the paper is organized as follows: in Section \ref{sec:related_works} we report the related works and a brief analysis of the state of the art, Section \ref{sec:proposed_method} describes our study with the proposed evaluation metric, and Section \ref{sec:results} discusses the obtained results. Finally, Section \ref{sec:conclusions} concludes our paper with some future work.

\section{Related Works}
\label{sec:related_works}
This section reports the state-of-the-art related works. In particular, Section \ref{subsec:sota_mde} describes ViT-based models for MDE, and Section \ref{subsec:sota_famv} reports input feature attribution techniques, focusing on the computer vision field. Finally, in Section \ref{subsec:sota_xai_mde}, we discuss the existing explainability methods for MDE.

\subsection{Monocular Depth Estimation}
\label{subsec:sota_mde}
MDE \cite{mde} is the task of predicting a dense depth map from a single RGB image, reconstructing the spatial distance of scene elements relative to the camera, recently achieved through deep neural networks.
In particular, some ViT architectures have emerged as effective solutions for this task \cite{mde_vit}. Notable examples include Binsformer \cite{binsformer}, which treats the MDE as a classification-regression-based task through the generation of adaptive bins, and PixelFormer \cite{pixelformer}, a full transformer model. Such architectures can achieve high accuracy at the cost of slow inference time, due to the quadratic cost of the attention mechanism \cite{metaformer}. Therefore, some attention module optimizations have been proposed to address this inefficiency, such as \cite{metaformer}, which replaces the attention with simpler token mixers (e.g. pooling operations) and \cite{pyramid_vit}, which reduces the size of values and keys to lower the computational cost. Based on such optimizations, recent architectures like METER \cite{meter} and its optimized versions \cite{meta_pyra_meter} have been developed to reduce the complexity of lightweight models and adapt them to hardware-constrained devices.

Despite the progress in model design, the explainability of MDE models, both deep and lightweight, remains an open challenge due to their black-box nature. To address this gap, we conduct our study on two mentioned literature models, i.e. PixelFormer and METER, to assess well-established explainability methods on deep and lightweight models.

\subsection{Input Feature Attribution Methods for Vision}
\label{subsec:sota_famv}
The explainability of computer vision models mainly focuses on input feature attribution methods, aiming to identify those features that mostly influence the model's output.
Notable algorithms are gradient-based, such as Saliency Maps \cite{saliency}, Integrated Gradients \cite{int_grad}, and GradCAM \cite{gradcam}, which rely on the gradient of the network to highlight its interesting regions. In particular, Saliency Maps compute the gradient of the model's output with respect to the input to determine which pixels of the input image follow the gradient direction. At the same time, Integrated Gradients sum the gradients along a path from a baseline. Another well-known method is Layer-wise Relevance BackPropagation (LRP) \cite{lrp}, whose goal is to explain each input feature’s contribution to the model’s output by assigning a relevance score to each neuron. Saliency Maps and Integrated Gradients are model-agnostic methods, meaning that they can be applied to any architecture, in contrast with LRP and GradCAM, which are model-specific and tailored for CNNs. In the context of model-specific techniques, Attention Rollout \cite{attn_roll} leverages the attention weights to aggregate information across the layers of a model, making it suitable for transformers. This method has been developed to improve the explicit attention visualization, consisting of taking the attention weights of a single layer at once, without considering the entire model architecture and leading to a loss of information flow in the network.

However, such techniques typically produce reliable results in classification tasks \cite{attn_roll} \cite{gradcam}, where discrete and localized outputs are often produced with a class token engaging the prediction information. On the other hand, their effectiveness in dense tasks such as MDE is far from obvious due to the spatially dependent relationships among the features distributed across the model. Building on this, in the following we investigate the application of Saliency Maps, Integrated Gradients, and Attention Rollout in explaining the input-output relation in neural networks designed for MDE.

\subsection{Explainability of MDE Models}
\label{subsec:sota_xai_mde}
Despite the increase of XAI in various applications to create trustworthiness and reliability towards deep learning models, explaining the decision-making process of neural networks for MDE is an open field of study \cite{mde_xai_1} \cite{mde_xai_2}. In the depth map prediction, each input pixel affects the prediction for other pixels, making it harder to localize contributions. Moreover, 
The sparseness of the feature information across the entire network makes studying explainability techniques harder on such models.
Few works in the literature try to generate reliable explanations for MDE, implementing interpretable networks as in \cite{mde_xai_1}. The authors of this work find that some neurons in the network are responsible for specific depth ranges, allowing the model to concentrate on objects near and far from the camera. Based on the mentioned observations, they quantify the interpretability of MDE models by assigning each neuron to its depth range. Moreover, they propose a technique to train interpretable MDE networks without modifying their structure. A different solution is proposed in \cite{mde_xai_2}, where the knowledge distillation leverages an explainability map from the teacher to allow for interpretable student learning. 

However, the mentioned works don't analyze a comparison between literature methods or quantify the importance of the highlighted input regions exclusively for the predicted depth map.

\section{Proposed Method}
\label{sec:proposed_method}
In this section, we present our explainability study of MDE models and the introduction of the novel AF metric to evaluate visual explanations.

\subsection{Explainability Study for MDE}
\label{subsec:expl_study_for_mde}
We first decided to apply three well-established explainability methods to highlight the most important input features of MDE models: Saliency Maps \cite{saliency}, Integrated Gradients \cite{int_grad}, and Attention Rollout \cite{attn_roll}. The first two techniques work with the gradient of the network, allowing one to understand how the gradient information affects the depth map prediction, while the third one leverages attention weights to get an overview of the interactions between the layers of the network. All are input feature attribution methods, meaning that they assign a relevance score for the output prediction to each input feature.

Formally, given our model $f$ and an input $x \in \mathbb{R}^d$, an attribution feature method is defined as the following map:

\begin{equation}
\label{eq:feat_attr}
    f(x), x \rightarrow \alpha_1, \alpha_2, \dots, \alpha_d
\end{equation}
where $\alpha_i$ is the relevance of the $i$-th input feature on the prediction.

According to \eqref{eq:feat_attr}, each input feature attribution method should be defined consequently.
Therefore, regarding Saliency Maps, we take the maximum of the gradient over the channels $c$ to attribute a relevance score to every input feature:

\begin{equation}
\label{eq:saliency_map}
    SaliencyMap(x) = \max_c(|\nabla f_x(x)|).
\end{equation}

While the Saliency Maps method \eqref{eq:saliency_map} computes the gradient just once, Integrated Gradients defines a path between the input image $x$ and a baseline (e.g. black) image $x'$ and computes the gradients at all points along the path with a number of steps $m$ in the following way \cite{int_grad}:

\begin{equation}
\label{eq:int_grad}
\begin{split}
    IntegratedGradients(x) = \\ (x \text{—} x') \times \sum_{k=1}^m \frac{\partial f(x' + \frac{k}{m} \times (x \text{—} x'))}{\partial x} \times \frac{1}{m},
\end{split}
\end{equation}



where $k = 1, \dots, m$.

Both for Saliency Maps and Integrated Gradients methodologies, we exploit the mean of all the depths contained in the output depth map to compute the gradient with respect to the input image. As can be noticed from \eqref{eq:int_grad}, the Integrated Gradients method calculates such a mean every time the gradient is computed.

While the above mentioned techniques exploit the gradient regardless of the task on which the model is tailored, the Attention Rollout method requires some modifications for MDE. In fact, in its original version, it leverages the class token associated with the network's prediction to attribute a score to each input feature. Because of the absence of a class token in MDE,  the information coming from the attention weights needs to be aggregated.

Specifically, we aggregate the attention matrices of the heads by using mathematical operations such as mean, minimum, or maximum \cite{attn_roll}, as expressed in  \eqref{eq:aggr}:

\begin{equation}
\label{eq:aggr}
    B^l = \psi(A^{(1,l)}, A^{(2,l)}, \dots, A^{H,l}),
\end{equation}
where $\psi(\cdot)$ denotes the aggregation operation (e.g. mean, min or max), $H$ is the number of heads, $l$ is the selected layer ($l=1,2,\dots, L$ and L is the number of layers)

Then we collect all the $B^l$ matrices over the layers \cite{attn_roll}:

\begin{equation}
\label{eq:layer_aggr}
    M^{(L-t-1)} = B_{res}^{(L-t-1)}M^{(L-i)},
\end{equation}

where $B_{res}^{(L-t-1)} = \frac{1}{2}B^{(L-t-1)}+\frac{1}{2}I$ and $t = 0,1,\dots,L-1$. To start, $M^L$ is initialized to $B^L$.

The final aggregated attention matrix $M^0 \in \mathbb{R}^{n \times m}$ resulting from \eqref{eq:layer_aggr} provides the relationships between input patches. Each entry $M^0_{ij}$ expresses the relative importance of patch $j$ to patch $i$. To compute the absolute importance of every input patch, we sum along the columns of $M^0$ into a single vector $s \in \mathbb{R}^{n}$:

\begin{equation}
\label{eq:vector}
    s_i=\sum_{j=1}^{m} M^0_{ij}, \text{ for } i = 1,2,\dots,n
\end{equation}

Equation \eqref{eq:vector} reports how to compute each entry of the vector $s$ to obtain the overall importance of each patch of the input image.






\subsection{Evaluation Framework}
\label{subsec:eval_framew}

\begin{figure*}[htbp]
\centerline{\includegraphics[width=\textwidth]{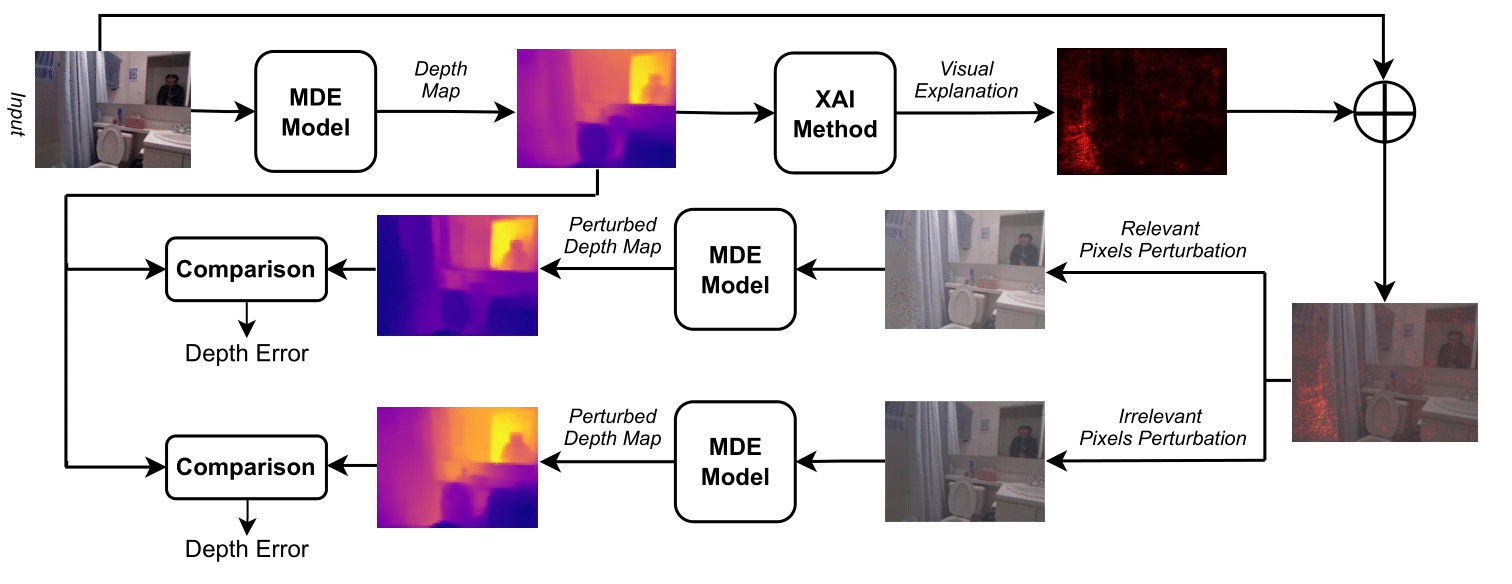}}
\caption{Pipeline of the exploited evaluation framework.}
\label{eval_framew_figure}
\end{figure*}

To comprehensively evaluate the validity of the visual maps generated by the explainability methods, we perturb the most relevant and most irrelevant pixels, according to the visual explanation, and measure the difference in the output prediction by feeding the model with the perturbed input versions \cite{faith_est}. The visual pipeline of the evaluation framework is depicted in Fig. \ref{eval_framew_figure}.

The procedure involves selecting two sets of pixels: the first with the highest relevance and the second with the lowest relevance according to the visual map. These pixels are then perturbed to create two distinct modified versions of the original input image.
Subsequently, such perturbed inputs are given to the model to obtain two perturbed depth maps. We analyze the contribution of the perturbed pixels by computing the distances between the perturbed depth maps and the original prediction obtained from the original input, finding the corresponding depth errors.

Therefore, the described evaluation framework validates the effectiveness of the visual explanations and, in general, the reliability of the explainability methods applied within the MDE context.


\subsection{Attribution Fidelity}
\label{subsec:af}
Assessing the visual maps generated by explainability methods is essential to study their effectiveness. In the MDE feature attribution context, evaluating explanations is not trivial given the complexity of mapping the input relevance to the predicted depth map. The existing metrics, such as RMSE and the Faithfulness Estimate (FE) \cite{faith_est} cannot provide reliable insights for MDE explainability in some cases. 
Specifically, the RMSE computes the distance between two images but, on its own, does not provide a comprehensive evaluation of the explanation. Similarly, the FE calculates the correlation between the relevance scores of the pixels assigned by the explainability methods and the error produced by perturbing these pixels in descending order of relevance score. The problem is that such a metric also considers intermediate pixels in terms of relevance score, while our focus is on perturbing the most and least relevant pixels.

In our MDE scenario, given the adopted framework described in Section \ref{subsec:eval_framew}, what we want ideally is the depth error between the image coming from the perturbation of relevant pixels to be higher than the depth error coming from the perturbation of irrelevant pixels.
The problem is that we would also like the distance between these two depth errors to be the highest positive possible, meaning that relevant pixels are actually more impactful than irrelevant ones. This scenario should be verified for all the images in the dataset.

To this end, we propose the novel AF metric to comprehensively evaluate all these aspects using a single approach. 

Mathematically, given a model $f$, the input image $x$, the image obtained from $x$ by perturbing the relevant pixels $x_{rel}$, and the image obtained by perturbing the irrelevant pixels $x_{irr}$, the AF metric is defined in the following way:

\begin{equation}
\label{eq:af}
    AF = \frac{|d(f(x_{rel}),f(x))|\text{—}|d(f(x_{irr}),f(x))|}{|d(f(x_{rel}),f(x))|+|d(f(x_{irr}),f(x))|} \in (-1, 1),
\end{equation}

where $d(\cdot)$ is a metric to compute the distances between the model predictions (i.e. the depth errors reported in Fig. \ref{eval_framew_figure}).

The AF metric assesses a visual explanation for MDE by considering the following fundamental aspects:

\begin{itemize}
    \item It rewards high and positive differences between the depth errors caused by $x_{rel}$ and $x_{irr}$, both compared with the original depth map $f(x)$.
    \item It rewards high depth errors caused by $x_{rel}$ and small depth errors caused by $x_{irr}$.
\end{itemize}

To facilitate its interpretation and usability, AF is normalized between $-1$ and $1$. When the value is close to $1$, the impact of relevant pixels is much higher than that of irrelevant pixels, suggesting that the explainability method is working appropriately. When close to $0$, the impact of relevant and irrelevant pixels is the same, meaning that the explainability method is not able to distinguish between them. Finally, when close to $-1$, the irrelevant pixels are more important to the prediction than the relevant ones, underlying that the method erroneously exchanges the attribution scores.

\section{Results}
\label{sec:results}
In this section, we report the results obtained by adopting the explainability methods described in Section \ref{subsec:expl_study_for_mde} and evaluating them as depicted in Section \ref{subsec:eval_framew}. Moreover, we test our novel metric proposed in Section \ref{subsec:af}.

\subsection{Experimental Setup}
All the algorithms have been implemented in Python by leveraging the PyTorch deep learning framework. For the configuration of the explainability methods parameters, the Integrated Gradients technique was executed with $m = 200$ and a black image as baseline. At the same time, the $min$ operation has been exploited as the $\psi(\cdot)$ operator in the Attention Rollout configuration \eqref{eq:aggr}. 

In terms of perturbation techniques, we adopted three different approaches: (1) applying a simple black mask on the pixel of interest, (2) introducing Gaussian noise with mean $\mu=0.0$ and standard deviation $\sigma=0.8$, (3) leveraging the FGSM adversarial attack \cite{fgsm} with a perturbation magnitude of 3.0.
These perturbation procedures were applied to three different percentages of pixels (i.e. $1\%, 5\%, 10\%$) to analyze how varying the number of perturbed pixels affects the behaviour of the explainability methods. 

Regarding the evaluation metrics, we use the RMSE to compute the depth error between the depth map predicted by perturbing the relevant pixels and the original prediction (rRMSE). The same applies to the depth error caused by the perturbation of irrelevant pixels (iRMSE). As a consequence, the RMSE has been used as $d(\cdot)$ operator in the AF formulation \eqref{eq:af}. To clarify, if the 5\% of pixels are perturbed, then the rRMSE comes from the perturbation of the 5\% of pixels with the highest relevance scores, while the iRMSE results from the ones with the lowest scores.
Additionally, the Attribution Success Rate (ASR) was computed for each method across the test set. The ASR metric increases when the rRMSE surpasses the iRMSE for a given image and is calculated as the ratio of the number of such images to the total number of images, yielding a value between $0$ and $1$. Moreover, we adopt the FE \cite{faith_est} that checks the correlation between the impact of the perturbed pixels and their relevance score assigned by the explainability methods. Its range is $(-1, 1)$. Finally, we measure the novel proposed AF metric.

We experiment with the described procedure on two state-of-the-art models trained with the procedures described in the corresponding papers: PixelFormer \cite{pixelformer} and METER \cite{meter}, both with their large versions trained on the NYU dataset \cite{nyu}. We conducted our experiments on the test set of such a dataset, composed of 654 images with the corresponding depth maps.

\subsection{Explainability Methods Performance}
\begin{table}[h!]
    \centering
    \small
    \renewcommand{\arraystretch}{0.9} 
    \setlength{\tabcolsep}{4pt}      
    \caption{Metrics for Explainability Methods (EM): SM for Saliency Maps, IG for Integrated Gradients, and AR for Attention Rollout evaluated on the METER model using three perturbation types (Pert) across three pixel percentages (\%). The best results are highlighted in bold. Grey rows indicate examples relevant to the AF analysis, though other rows may also contribute to the discussion.}
    \resizebox{\columnwidth}{!}{%
    \begin{tabular}{|l|c|c|c|c|c|c|c|}
        \hline
        \textbf{EM} & \textbf{Pert} & \textbf{\%} & \textbf{rRMSE} $\uparrow$ & \textbf{iRMSE} $\downarrow$ & \textbf{ASR} $\uparrow$ & \textbf{FE} $\uparrow$ & \textbf{AF} $\uparrow$ \\
        \hline
        \multirow{9}{*}{\textbf{SM}} 
        & \multirow{3}{*}{black} & 1 & 26.31 & 16.69 & 0.79 & 0.09 & 0.15 \\
        &                       & 5 & 39.23 & 24.69 & 0.76 & 0.14 & 0.15 \\
        &                       & 10 & 46.64 & 30.87 & 0.75 & 0.16 & 0.12 \\
        \cline{2-8}
        & \multirow{3}{*}{gauss} & \cellcolor{gray!20}1 & \cellcolor{gray!20}16.20 & \cellcolor{gray!20}\textbf{5.28} & \cellcolor{gray!20}\textbf{0.97} & \cellcolor{gray!20}0.05 & \cellcolor{gray!20}\textbf{0.49} \\
        &                       & \cellcolor{gray!20}5 & \cellcolor{gray!20}24.76 & \cellcolor{gray!20}12.65 & \cellcolor{gray!20}\textbf{0.97} & \cellcolor{gray!20}0.05 & \cellcolor{gray!20}0.32 \\
        &                       & 10 & 28.64 & 15.69 & 0.93 & 0.07 & 0.25 \\
        \cline{2-8}
        & \multirow{3}{*}{fgsm} & 1 & 143.02 & 173.70 & 0.38 & 0.23 & -0.08 \\
        &                       & 5 & 162.07 & 131.67 & 0.84 & 0.29 & 0.11 \\
        &                       & \cellcolor{gray!20}10 & \cellcolor{gray!20}167.21 & \cellcolor{gray!20}133.24 & \cellcolor{gray!20}0.89 & \cellcolor{gray!20}0.29 & \cellcolor{gray!20}0.12 \\
        \cline{2-8}
        \hline \hline

        \multirow{9}{*}{\textbf{IG}}
        & \multirow{3}{*}{black} & 1 & 22.12 & 10.14 & 0.83 & -0.06 & 0.29 \\
        &                       & \cellcolor{gray!20}5 & \cellcolor{gray!20}36.28 & \cellcolor{gray!20}23.08 & \cellcolor{gray!20}0.80 & \cellcolor{gray!20}-0.04 & \cellcolor{gray!20}0.16 \\
        &                       & 10 & 42.49 & 28.31 & 0.66 & 0.03 & 0.08 \\
        \cline{2-8}
        & \multirow{3}{*}{gauss} & 1 & 12.53 & 7.48 & 0.51 & -0.12 & 0.04 \\
        &                       & 5 & 22.12 & 15.01 & 0.61 & -0.05 & 0.06 \\
        &                       & 10 & 27.47 & 19.20 & 0.49 & 0.03 & 0.02 \\
        \cline{2-8}
        & \multirow{3}{*}{fgsm} & \cellcolor{gray!20}1 & \cellcolor{gray!20}129.98 & \cellcolor{gray!20}145.67 & \cellcolor{gray!20}0.41 & \cellcolor{gray!20}-0.62 & \cellcolor{gray!20}-0.06 \\
        &                       & 5 & 133.65 & 132.22 & 0.51 & -0.40 & 0.003 \\
        &                       & 10 & 138.24 & 114.05 & 0.75 & -0.47 & 0.09 \\
        \cline{2-8}
        \hline \hline

        \multirow{9}{*}{\textbf{AR}}
        & \multirow{3}{*}{black} & 1 & 18.08 & 11.41 & 0.61 & 0.31 & 0.07 \\
        &                       & 5 & 36.72 & 23.31 & 0.71 & 0.37 & 0.13 \\
        &                       & 10 & 47.75 & 29.59 & 0.72 & 0.38 & 0.12 \\
        \cline{2-8}
        & \multirow{3}{*}{gauss} & 1 & 11.89 & 8.27 & 0.52 & 0.23 & 0.02 \\
        &                       & 5 & 23.12 & 15.88 & 0.73 & \textbf{0.41} & 0.10 \\
        &                       & 10 & 83.03 & 76.84 & 0.56 & -0.20 & 0.05 \\
        \cline{2-8}
        & \multirow{3}{*}{fgsm} & 1 & 143.67 & 133.23 & 0.59 & -0.19 & 0.04 \\
        &                       & 5 & 165.69 & 146.68 & 0.58 & -0.08 & 0.07 \\
        &                       & 10 & 173.15 & 150.28 & 0.61 & -0.13 & 0.08 \\
        \cline{2-8}
        \hline
    \end{tabular}
    }
    \label{tab:meter_results}
\end{table}

\begin{table}[h!]
    \centering
    \small
    \renewcommand{\arraystretch}{0.9} 
    \setlength{\tabcolsep}{4pt}      
    \caption{Metrics for Explainability Methods (EM): SM for Saliency Maps, IG for Integrated Gradients, and AR for Attention Rollout evaluated on the PixelFormer model using three perturbation types (Pert) across three pixel percentages (\%). The best results are highlighted in bold. Grey rows indicate examples relevant to the AF analysis, though other rows may also contribute to the discussion.}
    \resizebox{\columnwidth}{!}{%
    \begin{tabular}{|l|c|c|c|c|c|c|c|}
        \hline
        \textbf{EM} & \textbf{Pert} & \textbf{\%} & \textbf{rRMSE} $\uparrow$ & \textbf{iRMSE} $\downarrow$ & \textbf{ASR} $\uparrow$ & \textbf{FE} $\uparrow$ & \textbf{AF} $\uparrow$ \\
        \hline
        \multirow{9}{*}{\textbf{SM}} 
        & \multirow{3}{*}{black} & 1 & 6.61 & 3.60 & 0.81 & 0.53 & 0.23 \\
        &                       & 5 & 12.38 & 6.11 & 0.95 & 0.49 & 0.29 \\
        &                       & 10 & 16.10 & 7.92 & 0.97 & 0.51 & 0.30 \\
        \cline{2-8}
        & \multirow{3}{*}{gauss} & 1 & 4.97 & 1.79 & 0.97 & 0.51 & 0.40 \\
        &                       & 5 & 9.13 & 3.92 & 0.97 & 0.50 & 0.35 \\
        &                       & 10 & 12.10 & 5.24 & 0.97 & 0.51 & 0.35 \\
        \cline{2-8}
        & \multirow{3}{*}{fgsm} & 1 & 37.26 & 33.06 & 0.71 & 0.55 & 0.06 \\
        &                       & 5 & 48.17 & 40.40 & 0.74 & 0.48 & 0.10 \\
        &                       & 10 & 54.66 & 46.60 & 0.73 & 0.52 & 0.09 \\
        \hline \hline

        \multirow{9}{*}{\textbf{IG}}
        & \multirow{3}{*}{black} & \cellcolor{gray!20}1 & \cellcolor{gray!20}7.40 & \cellcolor{gray!20}1.85 & \cellcolor{gray!20}\textbf{1.00} & \cellcolor{gray!20}0.88 & \cellcolor{gray!20}\textbf{0.56} \\
        &                       & 5 & 14.45 & 4.13 & \textbf{1.00} & 0.88 & 0.53 \\
        &                       & 10 & 20.04 & 5.89 & \textbf{1.00} & 0.88 & 0.51 \\
        \cline{2-8}
        & \multirow{3}{*}{gauss} & 1 & 4.62 & 2.29 & 0.94 & 0.87 & 0.30 \\
        &                       & 5 & 9.03 & 4.36 & 0.98 & 0.88 & 0.32 \\
        &                       & 10 & 12.38 & 5.91 & 0.94 & 0.88 & 0.32 \\
        \cline{2-8}
        & \multirow{3}{*}{fgsm} & 1 & 32.95 & 34.90 & 0.41 & 0.91 & -0.02 \\
        &                       & \cellcolor{gray!20}5 & \cellcolor{gray!20}39.92 & \cellcolor{gray!20}40.81 & \cellcolor{gray!20}0.43 & \cellcolor{gray!20}\textbf{0.93} & \cellcolor{gray!20}-0.01 \\
        &                       & 10 & 45.04 & 45.67 & 0.51 & 0.91 & -0.001 \\
        \hline \hline

        \multirow{9}{*}{\textbf{AR}}
        & \multirow{3}{*}{black} & 1 & 3.35 & 3.30 & 0.52 & -0.05 & -0.003 \\
        &                       & 5 & 9.30 & 8.60 & 0.51 & 0.08 & 0.003 \\
        &                       & 10 & 13.92 & 13.56 & 0.45 & 0.19 & -0.01 \\
        \cline{2-8}
        & \multirow{3}{*}{gauss} & 1 & 1.43 & \textbf{1.31} & 0.58 & -0.12 & 0.05 \\
        &                       & 5 & 3.30 & 3.04 & 0.60 & -0.001 & 0.05 \\
        &                       & 10 & 5.10 & 4.43 & 0.63 & -0.08 & 0.06 \\
        \cline{2-8}
        & \multirow{3}{*}{fgsm} & 1 & 32.59 & 32.81 & 0.45 & -0.05 & -0.01 \\
        &                       & 5 & 35.14 & 35.81 & 0.44 & 0.06 & -0.02 \\
        &                       & 10 & 39.37 & 39.29 & 0.51 & -0.02 & -0.002 \\
        \hline
    \end{tabular}
    }
    \label{tab:pixelformer_results}
\end{table}

The overall performance of the three studied explainability methods over the MDE task is discussed in this section. In particular, we report in Tables \ref{tab:meter_results} and \ref{tab:pixelformer_results} the obtained results respectively for METER and PixelFormer.

Regarding METER, the best results are achieved by the saliency maps, with the highest AF (0.49) and the highest ASR (0.97) when the Gaussian noise is applied to the 1$\%$ of pixels. The highest FE (0.41) is provided by the Attention Rollout and the Gaussian noise over the 5\% of the pixels. This suggests that, while the Saliency Maps identify better the most relevant and least relevant pixels, Attention Rollout has a smoother behaviour as it assigns a correct relevance score also to the intermediate pixels between the most and least relevant ones. 

Concerning PixelFormer, the best results are obtained by adopting the Integrated Gradients and the black mask to the 1$\%$ of pixels, with an ASR of 1.00 and an AF of 0.56. The best value for the FE (0.93) is always provided by the Integrated Gradients method, but with the FGSM attack on the 5$\%$ of pixels. However, there are high FE values for all the cases in which the Integrated Gradients technique is involved, meaning that it reliably classifies intermediate pixels in terms of relevance score. 

Summarizing, we can notice that for both METER and PixelFormer, the rRMSE and iRMSE values are much higher when the FGSM is applied, meaning that it destroys the depth much more than the other perturbation techniques. This is due to the nature of the FGSM, which considers the sign of the gradient to make a targeted attack. Regarding the explainability methods, the Saliency Maps provide valid explanations for METER, with an average ASR of 0.81 and AF of 0.18, while the best results on PixelFormer are achieved by the Integrated Gradients with an average ASR of 0.81, an AF of 0.28, and a FE of 0.89. Furthermore, it is generally observed that reducing the percentage of perturbed pixels with the black mask and Gaussian noise increases the performance of the explainability methods on METER but leads to a deterioration in performance on PixelFormer. On the other hand, when the FGSM is applied, the explainability methods work better with a large number of pixels, since gradient-based attacks directly rely on the gradient sign rather than perturbation magnitude.

\subsection{Attribution Fidelity Analysis}
In this section, we want to focus on those cases in which the existing metrics are not completely reliable in assessing the explainability methods in the MDE context. Such cases are important demonstrations of the effectiveness of the novel proposed AF metric.

As already anticipated in Section \ref{subsec:af}, if the ASR and the FE metric are high, then the method is supposed to work well. The limitation of the ASR is that it increases independently of the difference between the rRMSE and iRMSE. Indeed, it's reasonable that the explainability method works well if such a difference is as high as possible. On the other hand, the limitation of the FE is that it considers all the pixels of the input image during its computation, while we are just interested in the ones with the highest and lowest relevance scores.

Building on such limitations, referring to Table \ref{tab:meter_results}, by considering the FGSM applied to the 1\% of pixels highlighted by the Integrated Gradients method, we can notice that the AF (-0.06) suggests a bad behaviour of such a method, according to the other metrics. However, even if the ASR and the FE are the same (Saliency Maps combined with Gaussian noise applied to 1$\%$ and 5$\%$ of pixels), the AF is quite different, 0.49 for 1$\%$ and 0.32 for 5$\%$. This is because the AF also considers the magnitude of the depth errors, rewarding an iRMSE of 5.28 rather than 12.65. Additionally, in the case of Integrated Gradients combined with the black mask on the 5$\%$ of pixels, we can notice an ASR of 0.8, indicating an apparent good result of the explainability method. However, the value of 0.16 for the AF is quite low, suggesting the method is not working so accurately. This means that the pixels with the highest relevance scores are effectively more impactful than the ones with the least scores, but the difference is not so large.

\begin{figure}[ht]
    \centering
    \footnotesize rRMSE=40.17, iRMSE=8.49, FE=-0.04, AF=0.65 \\[3pt]
    \begin{minipage}[b]{0.05\linewidth} 
        \centering
        \raisebox{0.3\height}{\rotatebox{90}{\small METER}}
    \end{minipage}
    \hfill
    \begin{minipage}[b]{0.3\linewidth} 
        \centering
        \includegraphics[width=\linewidth]{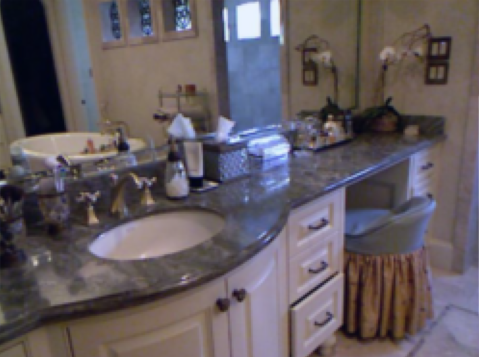}
        \vspace{-6mm}
        \caption*{\scriptsize Input Image}
    \end{minipage}
    \hfill
    \begin{minipage}[b]{0.3\linewidth}
        \centering
        \includegraphics[width=\linewidth]{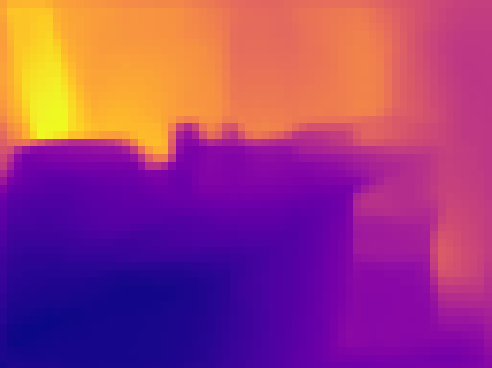}
        \vspace{-6mm}
        \caption*{\scriptsize Predicted Depth Map}
    \end{minipage}
    \hfill
    \begin{minipage}[b]{0.3\linewidth}
        \centering
        \includegraphics[width=\linewidth]{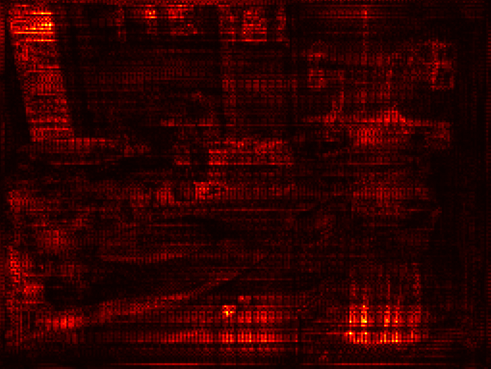}
        \vspace{-6mm}
        \caption*{\scriptsize Visual Explanation}
    \end{minipage}

    \vspace{0.2cm}

    \footnotesize rRMSE=24.59, iRMSE=21.63, FE=0.93, AF=0.06 \\[3pt]
    \begin{minipage}[b]{0.05\linewidth} 
        \centering
        \raisebox{0.2\height}{\rotatebox{90}{\small PixelFormer}}
    \end{minipage}
    \hfill
    \begin{minipage}[b]{0.3\linewidth}
        \centering
        \includegraphics[width=\linewidth]{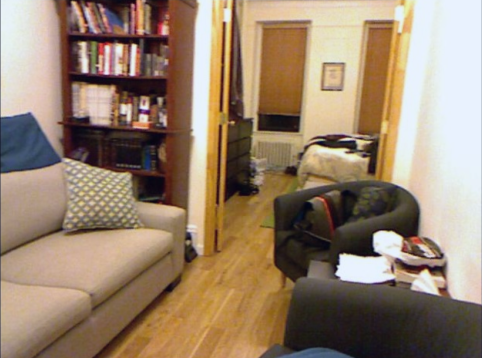}
        \vspace{-6mm}
        \caption*{\scriptsize Input Image}
    \end{minipage}
    \hfill
    \begin{minipage}[b]{0.3\linewidth}
        \centering
        \includegraphics[width=\linewidth]{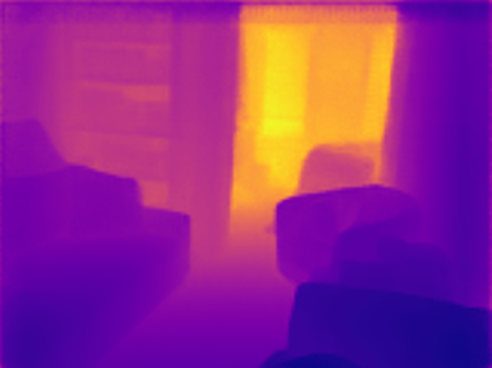}
        \vspace{-6mm}
        \caption*{\scriptsize Predicted Depth Map}
    \end{minipage}
    \hfill
    \begin{minipage}[b]{0.3\linewidth}
        \centering
        \includegraphics[width=\linewidth]{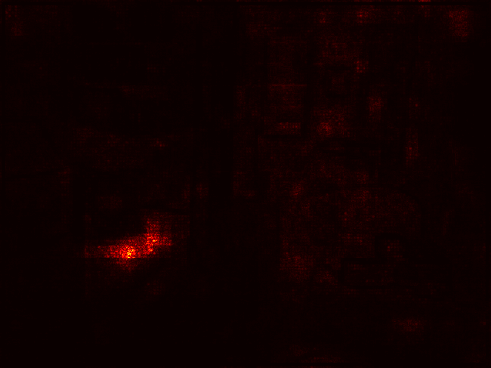}
        \vspace{-6mm}
        \caption*{\scriptsize Visual Explanation}
    \end{minipage}

    \caption{Visual explanations and metrics comparison for METER (visual map generated through Integrated Gradients, metrics computed with a black mask on the 5\% of pixels) and PixelFormer (visual map generated through Saliency Maps, metrics computed with FGSM on the 10\% of pixels). 
    }
    \label{fig:results}
\end{figure}

As for Table \ref{tab:pixelformer_results}, we can notice that when the Integrated Gradients are applied in combination with the FGSM on the 5$\%$ of pixels, the FE is extremely high (0.93) with ASR and AF values of 0.43 and -0.01, respectively. Such results highlight the limitation of the FE in evaluating also pixels with intermediate relevance scores (i.e. relevance scores between the highest and the lowest ones), while the AF confirms the weakness of the generated explanations.
Fig. \ref{fig:results} shows two examples of visual explanations along with the corresponding metrics. The explanation for METER reports an acceptable AF value given by the large difference between the rRMSE and iRMSE, despite a very low FE. 
In contrast, the visual explanation for PixelFormer highlights a very low AF value, even when the other metrics indicate favourable performance. 
Such examples illustrate the capacity of the AF metric to capture the validity of the visual explanations, even when other metrics could lead to conflicting assessments.

\section{Conclusions and future Works}
\label{sec:conclusions}
In this work, we address the challenge of studying and evaluating the explainability of MDE networks by adopting three well-established input feature attribution techniques, Saliency Maps, Integrated Gradients and Attention Rollout, on a deep full-transformer architecture, PixelFormer, and a lightweight ViT-based model, METER. We perturbed the highlighted pixels by the visual maps and fed the models with the obtained perturbed input to verify their contribution to the predicted depth map. Furthermore, due to the lack of reliable benchmarks to evaluate the obtained visual explanations in the MDE context, we propose a novel metric, AF, to reach a comprehensive assessment and compare various explainability methods. AF considers two fundamental aspects to analyze a visual map: the difference between the depth errors caused by the relevant and irrelevant pixel perturbations and the magnitude of the depth errors. Experimental results show that Saliency Maps provide reliable explanations for METER, while Integrated Gradients work well for PixelFormer. The AF metric demonstrates its reliability in identifying failures in explainability methods even when other metrics suggest acceptable performance, placing as a standard for evaluating MDE visual explanations. In the future, we plan to extend the MDE explainability study by increasing the range of attacks to perturb images and testing over other datasets. Moreover, we aim to analyze the AF metric using different explainability techniques for various tasks to reach a more comprehensive evaluation of the generated explanations.

\section*{Acknowledgments}
\noindent
This study has been partially supported by SERICS (PE00000014) under the MUR National Recovery and Resilience Plan funded by the European Union - NextGenerationEU.

\end{document}